\documentclass[sigconf]{acmart}
\usepackage{kotex}
\usepackage{amsmath}
\usepackage{booktabs}
\usepackage{multirow}
\usepackage{enumitem}
\usepackage{tabularx}
\usepackage{graphicx}    
\usepackage{subcaption}  
\usepackage{algorithm}
\usepackage{algpseudocode}

\theoremstyle{definition}
\newtheorem{definition}{Definition}

\AtBeginDocument{%
  }

\setcopyright{acmlicensed}
\copyrightyear{2018}
\acmYear{2018}
\acmDOI{XXXXXXX.XXXXXXX}
\acmConference[Conference acronym 'XX]{Make sure to enter the correct
  conference title from your rights confirmation email}{June 03--05,
  2018}{Woodstock, NY}
\acmISBN{978-1-4503-XXXX-X/2018/06}

\begin{document}
\title{ILLC: Iterative Layer-by-Layer Compression for Enhancing Structural Faithfulness in SpArX}

\author{Ungsik Kim}
\email{blpeng@gnu.ac.kr}
\orcid{1234-5678-9012}
\authornotemark[1]
\affiliation{%
  \institution{Gyeongsang National University}
  \city{Jinju}
  \state{Gyeongsangnam-do}
  \country{South Korea}
}

\author{Suwon Lee}
\affiliation{%
  \institution{Gyeongsang National University}
  \city{Jinju}
  \country{South Korea}}
\email{suwonlee@gnu.ac.kr}
\orcid{1234-5678-9012}
\authornotemark[2]


\renewcommand{\shortauthors}{Kim et al.}

\begin{abstract}
  In the field of Explainable Artificial Intelligence (XAI), argumentative XAI approaches have been proposed to represent the internal reasoning process of deep neural networks in a more transparent way by interpreting hidden nodes as arguements. However, as the number of layers increases, existing compression methods simplify all layers at once, which lead to high accumulative information loss. To compensate for this, we propose an iterative layer-by-layer compression technique in which each layer is compressed separately and the reduction error in the next layer is immediately compensated for, thereby improving the overall input-output and structural fidelity of the model. Experiments on the Breast Cancer Diagnosis dataset show that, compared to traditional compression, the method reduces input-output and structural unfaithfulness, and maintains a more consistent attack-support relationship in the Argumentative Explanation scheme. This is significant because it provides a new way to make complex MLP models more compact while still conveying their internal inference logic without distortion.
\end{abstract}




\maketitle

\section{Introduction}

In the field of explainable artificial intelligence (XAI), various attempts have been made to reveal the complex inner workings of deep learning models in a human-understandable form.\cite{doshi2017towards, guidotti2018survey} Especially in high-stakes domains such as medicine, autonomous driving and robotics, and finance, it is essential to interpret why a model produced a certain result.\cite{ayoobi2019handling} However, typical deep learning models have a large number of parameters and a deep layered structure for high predictive performance, creating a "black box" problem.\cite{arik2021tabnet}

Various methods have been proposed to reduce model complexity, such as weight pruning\cite{han2015learning} and knowledge distillation\cite{hinton2015distilling}, but these techniques do not follow the mechanism of the model, limiting the interpretability of the model output. In addition, the work of Argumentative XAI attempts to explain the internal reasoning process by analyzing the attack/support relationship of hidden nodes in a network by grouping them into arguments.\cite{vcyras2021argumentative, potyka2021interpreting} However, as deep learning models become increasingly large, reducing the complexity of the model comes at the cost of losing structural information or degrading predictive performance.

To address this trade-off between model fidelity and model simplicity, and to achieve higher explicability in the SparX framework, we propose a method to maximize model fidelity and structural similarity even under extreme compression of complex deep learning models. Specifically, by sequentially compressing the Multi-Layer Perceptron (MLP) structure layer by layer, we aim to preserve the interactions between hidden nodes intact, thus maintaining a similar behavior to the internal argument framework of the original model. The goal of this work is to provide an efficient compression technique that does not significantly degrade the final prediction performance, but also the interpretability of the model. The contributions of this paper are summarized below.

\begin{itemize}
  \item RQ: How can Iterative Layer-by-Layer Compression (ILLC) improve structural fidelity compared to traditional compression?
  \item Assumption: If the distribution of activation values of hidden nodes is successfully clustered in a particular layer, it can accurately represent the structural features of that layer, allowing the error introduced in the previous layer to be readjusted in the next layer.
\end{itemize}
Based on these research questions and assumptions, in this paper we validate the proposed method through experiments on the Breast Cancer Diagnosis Dataset\cite{breast_cancer} and explore the trade-off between model complexity and explanatory power.

Section~\ref{sec:related_work} examines representative approaches in Explainable Artificial Intelligence (XAI) and model compression, and reviews prior research specifically related to Argumentative XAI. Section~\ref{sec:preliminary} defines the fundamental concepts employed in this paper, while Section~\ref{sec:proposed_method} introduces the Iterative Layer-by-Layer Compression algorithm and explains the specific procedure for sequentially correcting errors occurring at each layer. In Section~\ref{sec:experiments}, the performance of the proposed method and Structural Unfaithfulness are quantitatively evaluated using the Breast Cancer diagnosis dataset\cite{breast_cancer}, and its effectiveness is examined from the perspective of argumentative explanation. Finally, Section~\ref{sec:conclusion} summarizes and discusses the overall results, and proposes potential synergistic effects between Argumentative XAI and model compression techniques, as well as directions for future research.

\section{Related Work}\label{sec:related_work}
\subsection{Explainable AI(XAI)}
In recent years, as the complexity of deep learning models has increased, interest in Explainable AI (XAI) has grown. Doshi-Velez and Kim \cite{doshi2017towards} emphasized that a rigorous definition and evaluation of explainability is necessary for model transparency and interpretability, and Guidotti et al. \cite{guidotti2018survey} systematically summarized various XAI approaches. These studies are broadly categorized into post-hoc and intrinsic approaches, with the former applying explanatory techniques after the fact to already trained models \cite{ribeiro2016should, lundberg2017unified}, and the latter embedding explanatory structures into the model training process itself \cite{alvarez2018towards, chen2019looks, agarwal2021neural, arik2021tabnet}.

\subsection{Argumentative XAI}
Argumentative XAI aims to express the "logical rationale" and "reasoning process" for model predictions in a human-understandable argument structure. Cyras et al. \cite{vcyras2021argumentative} proposed a framework that interprets the hidden nodes of an artificial neural network as arguments and connects the relationships between them as attack or support relationships to show what internal logic the model uses to make predictions. The framework uses the Quantitative Bipolar Argumentation Framework (QBAF)\cite{potyka2021interpreting}, which expresses the attack/support trunk lines as positive/negative weights, and provides an intuitive and structured explanation of the model's output.

Ayoobi et al. \cite{ayoobi2023sparx} proposed the Sparse Argumentative Explanations Framework, which compresses MLPs (reduced models) and then converts them into QAFs for explanation. On the other hand, ProtoArgNet \cite{ayoobi2023protoargnet} by the same authors applies the idea of prototypical networks inside CNNs to visualize and argue "prototypical" support regions for each class, thus increasing the interpretability of the Intrinsic method. In addition, Ayoobi et al. \cite{ayoobi2019handling} proposed a general-purpose argumentation-based architecture for autonomous recovery from unexpected failures in general-purpose robotic systems. Potyka et al \cite{potyka2023explaining} presented an argumentation structure for random forest models and proposed a way to simplify the complex internals of random forests using Markov Chains.

\subsection{Model Compression Methods}
To reduce model size while increasing explainability, techniques such as network pruning and knowledge distillation have been studied. techniques have been studied extensively. Han et al. \cite{han2015learning} showed how to significantly reduce the number of model parameters by using deep compression to remove unnecessary weights. Knowledge distillation by Hinton et al. \cite{hinton2015distilling} uses "soft targets" to transfer learning signals between large (teacher) and small (student) models, reducing model size while minimizing performance degradation. However, most pruning and distillation techniques focus on considering all weights at once, rather than layer-by-layer, or on removing only specific channels (e.g., channel pruning in CNN \cite{he2017channel}). For layer-by-layer compression, each layer can be simplified sequentially while retaining important expressive capabilities of the existing model. This has the advantage of {"compensating for losses in the previous layer in the very next layer"}. \cite{iandola2016squeezenet}.

\section{Preliminary}\label{sec:preliminary}

\begin{definition}[Multi-Layer Perceptron (MLP)]
  \label{def:mlp}
  \textit{An MLP} $\mathcal{M}$ is a tuple $(V, E, B, W, \varphi)$.
  $(V, E)$ is a directed graph.
  $V = \biguplus_{l=0}^{d+1} V_l$ consists of (ordered) layers of neurons;
  for $0 \le l \le d+1$, $V_l = \{\,v_{l,1}, \ldots, v_{l,|V_l|}\}$ is the set of neurons in layer $l$.
  We call $V_0$ the {input layer}, $V_{d+1}$ the {output layer}, and $V_l$ (for $1 \le l \le d$) the {$l$-th hidden layer}; $d$ is the {depth} of the MLP.
  $E \subseteq \bigcup_{l=0}^{d} (V_l \times V_{l+1})$ is a set of edges between adjacent layers; if $E = \bigcup_{l=0}^{d} (V_l \times V_{l+1})$, then the MLP is called {fully connected}.
  $B = \{b^1, \ldots, b^{d+1}\}$ is a set of bias vectors, where for $1 \le l \le d+1$, $b^l \in \mathbb{R}^{|V_l|}$.
  $W = \{W^0, \ldots, W^{d}\}$ is a set of weight matrices, where for $1 \le l \le d+1$, $W^l \in \mathbb{R}^{|V_{l+1}| \times |V_l|}$, such that $W^l_{j,i} = 0$ when $(v_{l+1,j}, v_{l,i}) \notin E$.
  $\varphi: \mathbb{R} \to \mathbb{R}$ is an {activation function}.
\end{definition}

Conceptually, base score assigns an apriori strength to the argument.
but, in MLP, the base score can't assign an apriori thing.
because, MLP's initial weights are random.
While there are many ways to initialize, such as standard gaussian, Xavier\cite{pmlr-v9-glorot10a}, He\cite{he2015delving}, etc,
not apriori.

\begin{definition}[Quantitative Argumentation Framework (QBAF)]
  \label{def:qbaf}
  A QBAF with domain $D \subseteq \mathbb{R}$ is a tuple $(A, E, B, w)$ that consists of
  \begin{itemize}
    \item $\mathcal{A} = \{a_1, \ldots, a_{|A|}\}$ is a set of arguments;
    \item $E \subseteq A \times A$ is a set of directed edges;
    \item $B: \mathcal{A} \to D$ is a base score;
    \item $w: \mathcal{A} \times \mathcal{A} \to \mathbb{R}$ is a weight function.
  \end{itemize}
  Edges with negative/positive weights are called {attack} and {support} edges,
  denoted by {Att}/{Sup}, respectively.
\end{definition}

In order to interpret the arguments in an edge-weighted
QBAF, we consider a modular semantics based on the relationship between QBAFs and MLPs noted earlier.\cite{potyka2021interpreting}

\begin{definition}[Aggregation and Influence Functions for MLP Modeling]
  \label{def:agg_forward}
  Consider a neural network $\mathcal{M}=(V,E,B,W,\varphi)$ with:
  Let $i$ be the index of the forward/backward iteration. Then:

  \smallskip
  \noindent
  \textbf{Forward Aggregation.}
  \[
    h_{a}^{(i+1)}
    \;:=\;
    \sum_{(b,a)\in E}\! w\bigl((b,a)\bigr)\,\cdot\,o_{b}^{(i)} + \varphi^{-1}\beta(a))
  \]
  where $h_{a}^{(i+1)}$ is the pre-activation of argument $a$ and $o_{b}^{(i)}$ is the output activation of argument $b$ at iteration~$i$, where $\varphi^{-1}\beta(a)$ is bias.

  \smallskip
  \noindent
  \textbf{Forward Influence.}
  \[
    o_{a}^{(i+1)}
    \;:=\;
    \varphi\bigl(h_{a}^{(i+1)}\bigr).
  \]
  This is the post-activation (output) of argument $a$.
  For integration with the QBAF model, we define the base strength $\beta(a)$ of an argument to be in the output space as well as the pre-activation space, and apply $\varphi^{-1}$ to get it into the pre-activation.
  Many activation functions(e.g. sigmoid, tanh) have inverse functions, and ReLU can define an inverse function by Ayoobi et al \cite{ayoobi2023sparx}.
  \[
    \varphi_{\mathrm{ReLU}}^{-1}(x) \;=\;
    \begin{cases}
      x, & \text{if } x > 0, \\
      0, & \text{otherwise}.
    \end{cases}
  \]

  \smallskip
  \noindent
  \textbf{Backward Aggregation.}
  For backpropagation, define
  \[
    g_{a}^{(i+1)}
    \;:=\;
    \sum_{(a,b)\in E} w\bigl((a,b)\bigr)\;\delta_{b}^{(i)},
  \]
  which collects the incoming gradients from successor arguments $b$.

  \smallskip
  \noindent
  \textbf{Backward Influence.}
  The gradient $\delta_{a}^{(i+1)}$ at argument $a$ is updated by
  \[
    \delta_{a}^{(i+1)}
    \;:=\;
    \begin{cases}
      \displaystyle
      \frac{\partial L}{\partial a} \cdot \varphi'\bigl(h_{a}^{(i+1)}\bigr),
       & \text{if $a$ is in the output layer}, \\[6pt]
      g_{a}^{(i+1)} \cdot \varphi'\bigl(h_{a}^{(i+1)}\bigr),
       & \text{otherwise}.
    \end{cases}
  \]
  Here, $L$ is the loss function (e.g.\ cross-entropy or MSE), and $h_{a}^{(i+1)}$ is the pre-activation of argument $a$.

  \smallskip
  \noindent
  \textbf{Convergence.}
  After iterating through all layers in terms of parameter update(training step), the final activation of argument $a$ can be written as
  \[
    o_{a}^{(\infty)}
    \;=\;
    \lim_{i\to\infty}\,o_{a}^{(i)}.
  \]

\end{definition}



\begin{definition}[Graphical Structure of Clustered MLP]
  Given an MLP $\mathcal{M}$ and a clustering $P = \uplus_{l=1}^d P_l$ of $\mathcal{M}$,
  the graphical structure of the corresponding clustered MLP $\mu$ is a directed graph
  $(V^\mu, E^\mu)$ with:
  \[
    V^\mu \;=\; \biguplus_{l=0}^{d+1} V_l^\mu
  \]
  consisting of (ordered) layers of cluster-neurons such that:
  \begin{itemize}
    \item The input layer $V_0^\mu$ consists of a singleton cluster-neuron $v_{\{0,i\}}$
          for every input neuron $v_i \in V_0$.

    \item The $l$-th hidden layer of $\mu$ (for $0 < l < d+1$) consists of a cluster-neuron $v_C$
          for every cluster $C \in P_l$.

    \item The output layer $V_{d+1}^\mu$ consists of a singleton cluster-neuron $v_{\{d+1,j\}}$
          for every output neuron $v_{d+1,j} \in V_{d+1}$.
  \end{itemize}
  \[
    E^\mu \;=\; \bigcup_{l=0}^{d} \bigl( V_l^\mu \times V_{l+1}^\mu \bigr).
  \]
\end{definition}

\begin{definition}[Parameters of Clustered MLP]
  Given an MLP $\mathcal{M}$, let $(V^M, E^M)$ be the graphical structure of the corresponding classical MLP $\mu$. Then for the cluster and edge aggregation functions $\mathrm{Agg}^b$ and $\mathrm{Agg}^e$, respectively, $\mu$ is
  \[
    (V^\mu, E^\mu, B^\mu, W^\mu, \varphi)
  \]
  with parameters $B^\mu, W^\mu$ as follows:
  \begin{itemize}
    \item For every cluster-neuron $v_C \in V^\mu$, the bias in $B^\mu$ of $v_C$ is $\mathrm{Agg}^b(C)$;
    \item For every edge $(v_{C1}, v_{C2}) \in E^\mu$, the weight in $W^\mu$ of the edge is $\mathrm{Agg}^e((C1, C2))$.
  \end{itemize}
\end{definition}

\section{Proposed Method}\label{sec:proposed_method}

\begin{figure*}[!htbp]
  \centering
  \includegraphics[width=1.0\textwidth]{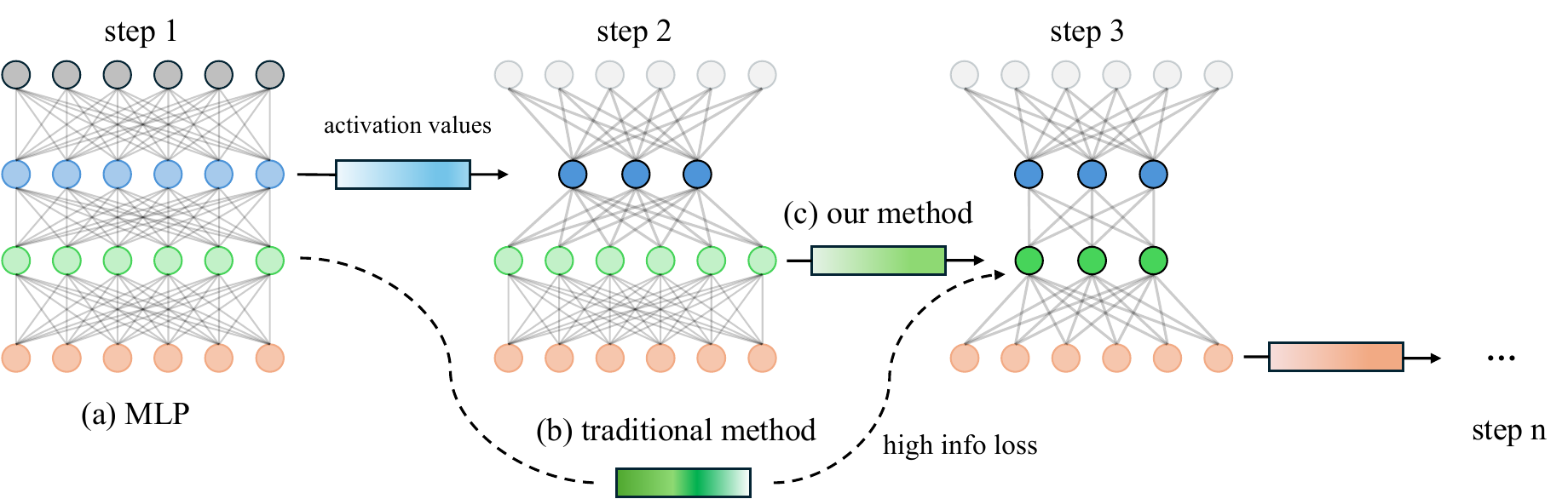}
  \caption{A visual illustration of how the "activation value updated in the previous layer" leads to the input of the next layer during Iterative Layer-by-Layer Compression (ILLC) to minimize error accumulation. (a) is the original MLP structure. In step 2, the original method and our method have the same behavior. (b) The original method compresses using activation values obtained by transferring from the original MLP model. (c) Our method compresses using activation values obtained by again forwarding from the model compressed in step 2. Since the output of layer 1, compressed and recalculated in step 1, becomes the input of step 2, the error in layer 2 compression can be corrected immediately. This maintains a finer activation pattern between layers than the existing method, which ultimately reduces both structural fidelity and input-output fidelity compared to the original model.}
  \label{fig:overview}
\end{figure*}

We propose \textbf{Iterative Layer-by-Layer Compression (ILLC)}, which compresses individual layers sequentially (iteratively) according to the activation distribution of each layer.(Algorithm~\ref{alg:illc}, Figure~\ref{fig:overview}) ILLC recalculates the activation values immediately after layer compression and immediately applies the result to the next layer to minimize the variation of activation values. This minimizes the loss of local activation distributions across layers and preserves the structural properties of the original model as much as possible.

\begin{algorithm}[!htbp]
  \caption{Iterative Layer-by-Layer Compression (ILLC)}
  \label{alg:illc}
  \begin{algorithmic}[1]
    \Require Original trained MLP model $\mathcal{M}$ with layers $V = \{V_0, V_1, \dots, V_{d+1}\}$
    \Require Dataset $X$, Shrinkage factor $\gamma \in (0,1]$
    \Ensure Compressed MLP model $\mu$

    \State Extract weight matrices and biases $\{W_l, b_l\}_{l=0}^d$ from $\mathcal{M}$

    \State Initialize compressed parameters $\{W_l^{\mu}, b_l^{\mu}\}_{l=0}^d \gets$ original parameters
    \State Set input data $X_{cur} \gets X$

    \For{layer $l = 1, 2, \dots, d$}
    \State Compute activations: $A_l = \sigma(X_{cur} \cdot W_{l-1} + b_{l-1})$
    \State Perform clustering on node activations $A_l^T$
    \State Set number of clusters $|C_l| = \gamma \cdot |V_l|$
    \State Obtain clustering labels $C_l$
    \For{each cluster $c$ in $C_l$}
    \State Merge nodes within cluster $c$:
    \State $W_{l-1, :, c}^{\mu} \gets \text{mean}(W_{l-1, :, \text{nodes}(c)}^{\mu})$
    \State $b_{l-1, c}^{\mu} \gets \text{mean}(b_{l-1, \text{nodes}(c)}^{\mu})$
    \State $W_{l, c,:}^{\mu} \gets \text{sum}(W_{l, \text{nodes}(c), :}^{\mu})$
    \EndFor
    \State Update $X_{cur} \gets \sigma(X_{cur} \cdot W_{l-1}^{\mu} + b_{l-1}^{\mu})$
    \EndFor

    \State Construct compressed MLP $\mu$ using updated parameters $\{W^{\mu}, b^{\mu}\}$
    \State \Return compressed model $\mu$
  \end{algorithmic}
\end{algorithm}

\subsection{Time Complexity of Compression}
Our method has a time complexity of $O(d)$, while the traditional method has a time complexity of $O(2d)$. If we were to compute the forward of the simple model over and over again, it would take $O(d^2)$, but we can reduce the redundant behavior by remembering the values before entering them into the activation function. The spatial complexity adds $O(N, |V_{l}|)$. There are many things that require a strict calculation of temporal complexity, such as clustering algorithms, but this is not a factor in the comparison.

\section{Experiments}\label{sec:experiments}
We used the BREAST CANCER dataset\cite{breast_cancer}. It consists of 569 samples and 30 numerical features. Each sample contains a set of morphological features extracted from the nuclei of cells in breast cancer tissue. Each sample is classified as either benign or malignant, and this binary classification problem is used to evaluate the accuracy of cancer diagnosis.

\subsection{Metrics}

\begin{definition}[Input-Output Unfaithfulness]
  \label{def:input_output_unfaithfulness}
  The local input-output unfaithfulness of $\mu$ to $\mathcal{M}$ with respect to input $x$ and dataset $\Delta$ is
  \[
    \mathcal{L}^{\mathcal{M}}(\mu)
    \;=\;
    \sum_{x' \in \Delta}
    \pi_{x',x}
    \;\sum_{v \in V_{d+1}}
    \Bigl(
    O_{x'}^{\mathcal{M}}(v)
    \;-\;
    O_{x'}^{\mu}(v)
    \Bigr)^{2}.
  \]
  The global input-output unfaithfulness of $\mu$ to $\mathcal{M}$ with respect to dataset $\Delta$ is
  \[
    \mathcal{G}^{\mathcal{M}}(\mu)
    \;=\;
    \sum_{x' \in \Delta}
    \;\sum_{v \in V_{d+1}}
    \Bigl(
    O_{x'}^{\mathcal{M}}(v)
    \;-\;
    O_{x'}^{\mu}(v)
    \Bigr)^{2}.
  \]
\end{definition}

\begin{definition}[Structural Unfaithfulness]
  \label{def:structural_unfaithfulness}
  Let $K_l$ be the number of clusters at hidden layer $l$ in $\mu$ ($0 \le l \le d$)
  and let $K_{d+1}$ be the number of output neurons.
  Let $K_{l,j}$ be the number of neurons in the $j$-th cluster-neuron $C_{l,j}$
  ($0 \le l \le d+1$, with $K_{d+1,j} = 1$).
  The {local structural unfaithfulness} of $\mu$ to $\mathcal{M}$
  with respect to input $x'$ and dataset $\Delta$ is defined as:
  \[
    \mathcal{L}^s(\mu)
    = \sum_{x' \in \Delta} \pi_{x',x}
    \sum_{l=1}^{L_1}
    \sum_{j=1}^{K_l}
    \bigl(
    O_{\mathcal{M}}^\mu (v_{l,i})
    \;-\;
    O_{\mathcal{M}}^\mu (C_{l,j})
    \bigr)^2.
  \]
\end{definition}
Structural Unfaithfulness is a measure of how well the activation values of each hidden node in the original network (or the functions it was responsible for) were "merged" into the compressed nodes.

\begin{definition}[Cognitive Complexity]
  Let $K_l$ be the number of clusters at hidden layer $l$ in $\mu$ ($0 < l \le d$).
  Then, the {cognitive complexity} of $\mu$ is defined as
  \[
    \Omega(\mu) \;=\; \prod_{0 < l \le d+1} K_l.
  \]
\end{definition}

Cognitive complexity has traditionally been used as a metric to measure the comprehensibility of software modules.\cite{campbell2018cognitive} We define cognitive complexity as a metric to measure the complexity of a model. The more complex the model, the higher the value. Keeping more clusters at each level increases the value exponentially, which is an indicator of how many intermediate nodes humans need to see to interpret the explanation. There is a trade-off between cognitive complexity and fidelity, and a good explanation requires high performance on both metrics.

\subsection{Global Explanation}
Global Explanation is used to analyze how a model generally behaves across a data set and how faithfully its internal logical structure is maintained. In other words, Global Explanation is designed to look beyond individual data points to assess the structural similarity of the model's behavior and predictive rationale across the entire distribution of input data.

In this study, based on the definition~\ref{def:global-agg}, Global Input-Output Unfaithfulness and Structural Unfaithfulness measure how accurately the compressed model preserves the input-output relationships and structural activation value patterns of the original model using the full dataset.

Table~\ref{tab:global_explanation} compares the Global Explanation performance between the original model and the proposed ILLC method (ours). The most striking observation in this experiment is that, across all combinations of architectures (number of layers: 5, 10, 20) and neurons (100, 200, 500), ILLC improves both the Input-Output Unfaithfulness and Structural Unfaithfulness compared to the conventional global compression approach. For example, in an MLP with 20 hidden layers and 100 neurons per layer, the Input-Output Unfaithfulness improves from 0.0131 with the original compression method to 0.0104 with the proposed ILLC method, a 26\% improvement. Moreover, the Structural Unfaithfulness is also reduced, from 0.4342 to 0.4084, which is approximately a 6\% improvement.

These results demonstrate that ILLC maintains a finer-grained pattern of activity between layers and actively minimizes the information loss that can occur with global compression.

\begin{figure*}[!htbp]
  \centering
  \begin{subfigure}[t]{0.33\linewidth}
    \centering
    \includegraphics[width=\textwidth]{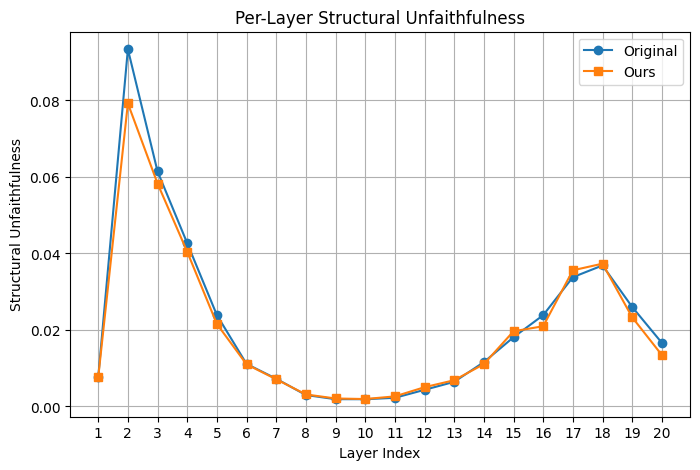}
    \caption{Structural Unfaithfulness by layer}
    \label{fig:sub1}
  \end{subfigure}
  \hfill
  \begin{subfigure}[t]{0.33\linewidth}
    \centering
    \includegraphics[width=\textwidth]{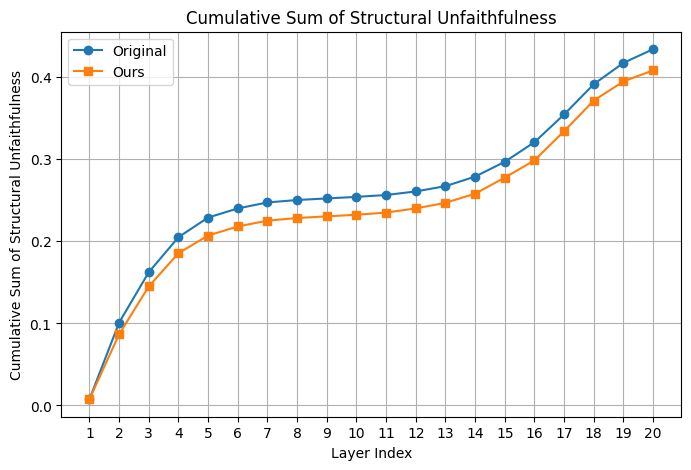}
    \caption{Cumulative Sum of Structural Unfaithfulness according to layer}
    \label{fig:sub2}
  \end{subfigure}
  \hfill
  \begin{subfigure}[t]{0.33\linewidth}
    \centering
    \includegraphics[width=\textwidth]{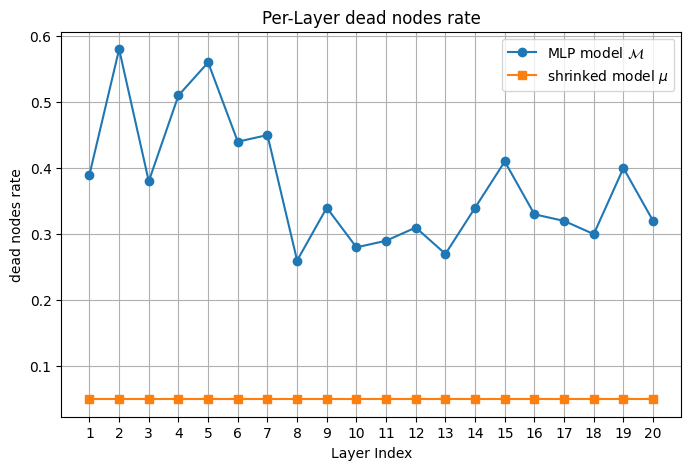}
    \caption{dead neurons ratio by layer}
    \label{fig:sub3}
  \end{subfigure}

  \caption{Analysis of layer 20 and 100 neurons MLP in Table~\ref{tab:global_explanation}}
  \label{fig:three_subfigs_fullwidth}
\end{figure*}

Figures \ref{fig:sub1} and \ref{fig:sub2} present the structural unfaithfulness graphs, indicating that the ILLC method tends to preserve each layer’s activation pattern in closer resemblance to the original model. This observation suggests that the hidden neurons—which play a critical role in the original model—continue to maintain a clear argumentative structure even after compression via ILLC. In particular, Figure \ref{fig:sub1} reveals a phenomenon termed Bi-Local-Maxima, where structural unfaithfulness experiences two pronounced increases in a specific layer. This effect arises during the sequential layer compression process: the loss in activation values in one layer is subsequently amplified in the following layer, as the slightly altered post-compression activation distribution is transmitted as input, thereby inducing a rebound effect. Similarly, the cumulative error graph in Figure \ref{fig:sub2} shows a sharp increase in structural error at certain layers due to the Bi-Local-Maxima effect; however, overall, ILLC maintains a significantly lower cumulative error growth rate compared to conventional methods.

Figure \ref{fig:sub3} illustrates the proportion of Dead Neurons that occur during the compression process. Dead Neurons refer to the phenomenon where an activation function (such as ReLU) consistently fixes a neuron’s output near zero, thereby reducing the model’s representational capacity. Traditional compression methods, which cluster without differentiating between the activation differences of Dead Neurons and normally active neurons, can lead to significant errors in the overall model structure. In light of the Bi-Local-Maxima phenomenon, layers with a high occurrence of Dead Neurons tend to exhibit more severe post-compression activation distortions, resulting in a pronounced rebound effect in the subsequent layer. In contrast, the ILLC method promptly corrects for these structural activation losses induced by Dead Neurons, thereby helping to minimize structural unfaithfulness.

\begin{definition}[Global Aggregation Functions]
  \label{def:global-agg}
  The average bias and edge aggregation functions are, respectively:
  \begin{align}
    \mathrm{Agg}^b(C) & = \frac{1}{|C|} \sum_{v_{1,i} \in C} b_i^l, \\[6pt]
    \mathrm{Agg}^e\bigl(C_1, C_2\bigr)
                      & = \sum_{v_{1,i} \in C_1} \frac{1}{|C_2|}
    \sum_{v_{1,j} \in C_2} W_{j,i}^l.
  \end{align}
\end{definition}

\begin{table}[!htbp]
  \begin{tabular}{@{}cccccc@{}}
    \toprule
    \multirow{2}{*}{Layer} & \multirow{2}{*}{Metric}       & \multirow{2}{*}{Method} & \multicolumn{3}{c}{Neurons per Layer}                                     \\ \cmidrule(l){4-6}
                           &                               &                         & 100                                   & 200             & 500             \\ \midrule
    \multirow{4}{*}{5}     & \multirow{2}{*}{input-output} & original                & 0.1811                                & 0.0071          & 0.0014          \\
                           &                               & ours                    & \textbf{0.1666}                       & 0.0071          & 0.0014          \\ \cmidrule(l){2-6}
                           & \multirow{2}{*}{Structural}   & original                & 1.4727                                & 0.5021          & 0.3335          \\
                           &                               & ours                    & \textbf{1.3671}                       & \textbf{0.4843} & \textbf{0.3327} \\ \midrule
    \multirow{4}{*}{10}    & \multirow{2}{*}{input-output} & original                & 0.0337                                & 0.0013          & 0.0008          \\
                           &                               & ours                    & \textbf{0.0137}                       & \textbf{0.0010} & 0.0008          \\ \cmidrule(l){2-6}
                           & \multirow{2}{*}{Structural}   & original                & 0.8394                                & 0.3122          & 0.1224          \\
                           &                               & ours                    & \textbf{0.7825}                       & \textbf{0.2968} & \textbf{0.1110} \\ \midrule
    \multirow{4}{*}{20}    & \multirow{2}{*}{input-output} & original                & 0.0131                                & 0.0014          & 0.0006          \\
                           &                               & ours                    & \textbf{0.0104}                       & \textbf{0.0009} & \textbf{0.0005} \\ \cmidrule(l){2-6}
                           & \multirow{2}{*}{Structural}   & original                & 0.4342                                & 0.3694          & 0.2621          \\
                           &                               & ours                    & \textbf{0.4084}                       & \textbf{0.3607} & \textbf{0.2604} \\ \bottomrule
  \end{tabular}
  \caption{Global Explanation of MLPs by each number of layers with compression rate 80\%}
  \label{tab:global_explanation}
\end{table}

\subsection{Local Explanation}
Local explanation is intended to interpret the model's prediction results and internal activation patterns with respect to the input of individual data samples. Through local explanations, one can analyze the reasons behind specific decisions at the level of individual inputs, which is particularly useful for explaining the diagnostic outcomes or predictive reasoning of individual cases.

In this experiment, we derived a compressed MLP from the original MLP based on Definition~\ref{def:local-edge-agg} and measured Local Unfaithfulness by comparing the activation values for each individual data sample \(x\) according to \ref{def:input_output_unfaithfulness}. Furthermore, Structural Unfaithfulness was assessed by comparing the activation values for each \(x\) according to \ref{def:structural_unfaithfulness}. These evaluations allowed us to assess the structural similarity between the two models and the reliability of the prediction outcomes for individual cases.

As shown in Table~\ref{tab:local_explanation}, in most instances the proposed method (ILLC) outperforms the conventional global compression (original) in both Local Input-Output Unfaithfulness and Structural Unfaithfulness metrics. For example, in the case of an MLP with 10 layers and 100 neurons per layer, the Input-Output Unfaithfulness improved from 0.0079 in the original model to 0.0044 with the ILLC approach—an improvement of approximately 79\%. Moreover, Structural Unfaithfulness showed a slight improvement from 0.5042 in the original model to 0.4943 with ILLC. These results indicate that when the activation value distributions of individual input data are more dramatically compressed in each layer, the detailed activation patterns inherent to the original model can be maintained with greater fidelity.

Specifically, as shown in Table~\ref{tab:local_explanation}, while the absolute value of Input-Output Unfaithfulness decreases with an increasing number of layers and neurons, Structural Unfaithfulness remains relatively constant. This observation suggests that as the model size increases, the intricate activation patterns become more complex, and minor discrepancies in these patterns may accumulate when using conventional compression methods. However, ILLC manages these subtle activation patterns on a per-layer basis, thereby better preserving the differences in activation patterns relative to the original model, regardless of the overall model size.

\begin{definition}[Local Edge Aggregation Function]
  \label{def:local-edge-agg}
  The local edge aggregation function with respect to the input $x$ is defined as:
  \begin{equation}
    \mathrm{Agg}^{e}(C_1, C_2)=
    \sum_{x' \in \Delta'} \pi_{x'}
    \sum_{v_{1,i} \in C_1}
    \frac{1}{|C_2| \cdot O_{x'}^{\mu}(v_{1,i})}
    \sum_{v_{1,j} \in C_2}
    W_{i,j}^\ell \, O_{x'}^{M}(v_{1,i})
  \end{equation}

  where $O_{x'}^{M}(v_{1,i})$ is the activation value of neuron $v_{1,i}$ in the original MLP,
  and $O_{x'}^{\mu}(C_1)$ is the activation value of the cluster-neuron $C_1$ in the clustered MLP.
  and $\pi_{x',x} = \exp\bigl(-\frac{D(x', x)^2}{\sigma^2}\bigr)$ is lower weight to a sample $x' \in \Delta'$ that is far from $x$.
\end{definition}

\begin{table}[!htbp]
  \begin{tabular}{@{}cccccc@{}}
    \toprule
    \multirow{2}{*}{Layer} & \multirow{2}{*}{Metric}       & \multirow{2}{*}{Method} & \multicolumn{3}{c}{Neurons per layer}                                     \\ \cmidrule(l){4-6}
                           &                               &                         & 100                                   & 200             & 500             \\ \midrule
    \multirow{4}{*}{5}     & \multirow{2}{*}{input-output} & original                & 0.0252                                & \textbf{0.0332} & 0.0005          \\
                           &                               & ours                    & \textbf{0.0220}                       & 0.0380          & \textbf{0.0004} \\ \cmidrule(l){2-6}
                           & \multirow{2}{*}{Structural}   & original                & 1.0155                                & 1.4132          & 0.3530          \\
                           &                               & ours                    & \textbf{0.9154}                       & \textbf{1.3469} & \textbf{0.3028} \\ \midrule
    \multirow{4}{*}{10}    & \multirow{2}{*}{input-output} & original                & 0.0079                                & 0.0010          & 0.0003          \\
                           &                               & ours                    & \textbf{0.0044}                       & \textbf{0.0010} & 0.0003          \\ \cmidrule(l){2-6}
                           & \multirow{2}{*}{Structural}   & original                & 0.5042                                & 0.2563          & \textbf{0.1350} \\
                           &                               & ours                    & \textbf{0.4943}                       & 0.2563          & 0.1351          \\ \midrule
    \multirow{4}{*}{20}    & \multirow{2}{*}{input-output} & original                & 0.0172                                & 0.0089          & 0.0004          \\
                           &                               & ours                    & \textbf{0.0152}                       & \textbf{0.0081} & 0.0004          \\ \cmidrule(l){2-6}
                           & \multirow{2}{*}{Structural}   & original                & \textbf{0.5459}                       & 0.6416          & 0.2551          \\
                           &                               & ours                    & 0.5517                                & \textbf{0.6354} & \textbf{0.2539} \\ \bottomrule
  \end{tabular}
  \caption{Local Explanation of MLPs with each number of layers by compression rate 80\%}
  \label{tab:local_explanation}
\end{table}

\section{Conclusion}\label{sec:conclusion}
In this paper, we propose a method called \textbf{ILLC} that performs sequential compression on a per-layer basis, in contrast to conventional clustering-based compression methods. Specifically, for the \(l\)-th layer, the activation matrix estimated from dataset \(D\) is partitioned into \(k_l\) clusters, based on which the compressed layer \(\mathcal{M}_l\) is constructed. Thereafter, the activation output of this layer is immediately incorporated as the input to the \((l+1)\)-th layer, and the weights and biases are recalculated to prevent the error from one layer from accumulating in the next. By proceeding sequentially from \(l=1\) to \(d\), the overall model structure is simplified while the unfaithfulness relative to the original model is minimized, resulting in the final model \(\mu\).

As demonstrated by the experimental results (Table~\ref{tab:global_explanation} and Table~\ref{tab:local_explanation}), employing the proposed iterative approach yields improvements in both input-output unfaithfulness and structural unfaithfulness compared to conventional methods. The argument explanation of the compressed model was not included because it remains unchanged from the existing method \cite{ayoobi2023sparx}.

Generally, fully connected layers are seldom increased in number due to the vanishing gradient problem---a challenge that Kaiming and Zhang addressed using residual learning \cite{he2016deep}. As future work, we aim to develop methods for explaining skip connections. When skip connections are present, clustering becomes more complex because it must be performed over the entire parallel branch rather than simply following the sequential order of layers.

Furthermore, applying ILLC to architectures other than MLPs may be a promising avenue for subsequent research. For instance, one could investigate how the clustering method might differ when applied to Graph Neural Networks.

Since the current approach relies solely on the activation values of hidden neurons, additional processing may be necessary when specialized activation functions (e.g., ReLU6 \cite{zou2020ship}) or normalization layers (e.g., BatchNorm) are involved.

The present study improves unfaithfulness while maintaining cognitive complexity. In future work, we plan to explore methods to enhance cognitive complexity, such as reducing the model's depth based on the universal approximation theorem \cite{cybenko1989approximation}.
\begin{acks}
  To Robert, for the bagels and explaining CMYK and color spaces.
\end{acks}



\bibliographystyle{ACM-Reference-Format}
\bibliography{sample-base}
\end{document}